\documentclass[10pt,twocolumn,letterpaper]{article}

\usepackage{cvpr}
\usepackage{times}
\usepackage{epsfig}
\usepackage{graphicx}
\usepackage{amsmath}
\usepackage{amssymb}
\usepackage{comment}
\usepackage{booktabs}
\usepackage{balance}

\def\sub#1{_{\rm #1}}
\def\vct#1{\mbox{\boldmath $#1$}}
\def\eg{{\it e.g.}}

\def\ie{{\it i.e.}}
\def\etal{{\it et al. }}

\usepackage[pagebackref=true,breaklinks=true,letterpaper=true,colorlinks,bookmarks=true]{hyperref}

\cvprfinalcopy 

\def\cvprPaperID{2895} 

\begin{document}

\title{Future Person Localization in First-Person Videos}

\author{Takuma Yagi\\
The University of Tokyo\\
Tokyo, Japan\\
{\tt\small tyagi@iis.u-tokyo.ac.jp}
\and
Karttikeya Mangalam\\
Indian Institute of Technology\\
Kanpur, India\\
{\tt\small mangalam@iitk.ac.in}\\ \\
\and
Ryo Yonetani\\
The University of Tokyo\\
Tokyo, Japan\\
{\tt\small yonetani@iis.u-tokyo.ac.jp} \\
\and
Yoichi Sato\\
The University of Tokyo\\
Tokyo, Japan\\
{\tt\small ysato@iis.u-tokyo.ac.jp}
}

\maketitle

\begin{abstract}
We present a new task that predicts future locations of people observed in first-person videos. Consider a first-person video stream continuously recorded by a wearable camera. Given a short clip of a person that is extracted from the complete stream, we aim to predict that person's location in future frames. To facilitate this future person localization ability, we make the following three key observations: a) First-person videos typically involve significant ego-motion which greatly affects the location of the target person in future frames; b) Scales of the target person act as a salient cue to estimate a perspective effect in first-person videos; c) First-person videos often capture people up-close, making it easier to leverage target poses (\eg, where they look) for predicting their future locations. We incorporate these three observations into a prediction framework with a multi-stream convolution-deconvolution architecture. Experimental results reveal our method to be effective on our new dataset as well as on a public social interaction dataset.
\end{abstract}

\section{Introduction}
\label{sec:introduction}

Assistive technologies are attracting increasing attention as a promising application of \emph{first-person vision} --- computer vision using wearable cameras such as Google Glass and GoPro HERO. Much like how we use our eyes, first-person vision techniques can act as an artificial visual system that perceives the world around camera wearers and assist them to decide on what to do next. Recent work has focused on a variety of assistive technologies such as blind navigation~\cite{Leung2014,Tian2013}, object echo-location~\cite{Tang2014}, and personalized object recognition~\cite{Kacorri2017}.

\begin{figure}[t]
  \begin{center}
    \includegraphics[width=\linewidth]{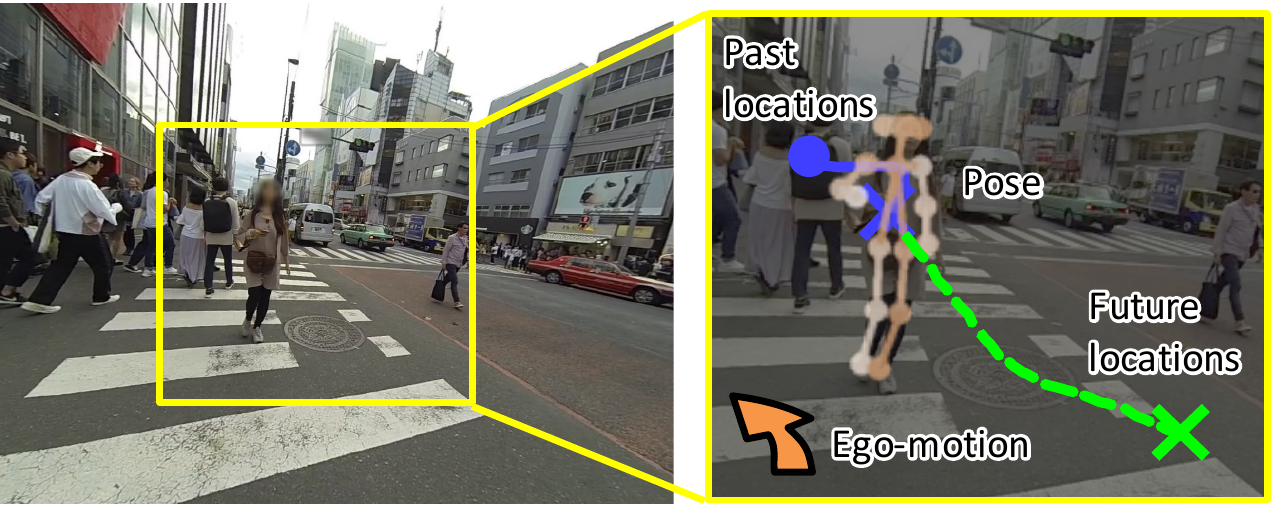}
    \caption{{\bf Future Person Localization}. Given a first-person video of a certain target person, our network predicts where the target person will be located in the future frames based on the poses and scales of the person as well as the ego-motions of the camera wearer.}
    \label{fig:teaser}
  \end{center}
\end{figure}

In this work, we are particularly interested in helping a user to navigate in crowded places with many people present in the user's vicinity. Consider a first-person video stream that a user records with a wearable camera. By observing people in certain frames and predicting how they move subsequently, we would be able to guide the user to avoid collisions. As the first step to realizing such safe navigation technologies in a crowded place, this work proposes a new task that predicts locations of people in future frames, \ie, \emph{future person localization}, in first-person videos as illustrated in Figure~\ref{fig:teaser}\footnote{Parts of faces in the paper were blurred for preserving privacy.}.

In order to enable future person localization, this work makes three key observations. First, \emph{ego-motion} of a camera wearer is clearly observed in the form of global motion of first-person videos. This ego-motion should be incorporated in the prediction framework as it greatly affects future locations of people. For example, if a camera wearer is moving forward, apparent vertical locations of people in the first-person video will be moving down accordingly. Moreover, if the camera wearer is walking towards people would change walking direction slightly to avoid a collision. This type of interacting behaviors would also affect the future locations of people.

Another key observation is that the scale of people acts as a salient cue to capture a perspective effect in first-person videos. Since the optical axis of a wearable camera tends to be parallel to the ground plane, visual distances in first-person video frames correspond to different physical distances depending on where people are observed in the frames. Such differences have to be taken into account for better future localization, especially when localizing people who are moving towards or away from the camera wearer.

The last key observation that improves the prediction capability is that, the pose of a person indicates how that person is moving and will be located in the near future. First-person videos can be used effectively to get access to such pose information as they often capture people up-close.

Based on these key observations, we propose a method to predict the future locations of a person seen in a first-person video based on ego-motions of the video, poses, scales, and locations of the person in the present and past video frames (also refer to Figure~\ref{fig:teaser}). Specifically, we develop a deep neural network that learns the history of the above cues in several previous frames and predicts locations of the target person in the subsequent future frames. A convolution-deconvolution architecture is introduced to encode and decode temporal evolution in these histories.

To validate our approach, we develop a new dataset of first-person videos called First-Person Locomotion (FPL) Dataset. The FPL Dataset contains about 5,000 people seen at diverse places. We demonstrate that our method successfully predicts future locations of people in first-person videos where state-of-the-art methods for human trajectory prediction using a static camera such as \cite{alahi2016social} fail. We also confirmed a promising performance of our method on a public first-person video dataset~\cite{fathi2012social}.

\section{Related Work}
A typical problem setting involving first-person vision is to recognize activities of camera wearers. Recently, some work has focused on activity recognition~\cite{Fathi2011,li2015delving,Ma_2016_CVPR,Pirsiavash2012}, activity forecasting~\cite{fan2017forecasting,Furnari:2017:NPE:3163595.3163822,ParkCVPR16,Rhinehart2017}, person identification~\cite{Hoshen2016}, gaze anticipation~\cite{Zhang_2017_CVPR} and grasp recognition~\cite{Bambach2015,Cai2015,li2013pixel,Saran2015}. Similar to our setting, other work has also tried to recognize behaviors of other people observed in first-person videos, \eg, group discovery~\cite{Alletto2015}, eye contact detection~\cite{Ye2015} and activity recognition~\cite{Ryoo:2015:RAP:2696454.2696462,Ryoo_2013_CVPR,yonetani2016recognizing}.

To the best of our knowledge, this work is the first to address the task of predicting future locations of people in first-person videos. Our task is different from \emph{egocentric future localization}~\cite{ParkCVPR16} that predicts where `the camera wearers' will be located in future frames. One notable exception is the recent work by Su~\etal~\cite{Su_2017_CVPR}. Although they proposed a method to predict future behaviors of basketball players in first-person videos, their method requires \emph{multiple} first-person videos to be recorded collectively and synchronously to reconstruct accurate 3D configurations of camera wearers. This requirement of multiple cameras is in contrast to our work (\ie, using a single camera) and not fit for assistive scenarios where no one but the user on assistance is expected to wear a camera.

Finally, the task of predicting future locations of people itself has been studied actively in computer vision. Given both locations of start and destination, work based on inverse reinforcement learning can forecast in-between paths~\cite{kitani2012activity,Ma_2017_CVPR}. Several methods have made use of Bayesian approaches~\cite{kooij2014context, Schneider2013}, recurrent neural networks~\cite{alahi2016social,lee2017desire}, fully-convolutional networks~\cite{Huang2016,yi2016pedestrian}, and other social or contextual features~\cite{Robicquet2016,Xie2013} for predicting human trajectories from images or videos. These methods are, however, not designed to deal with first-person videos where significant ego-motion affects the future location of a certain person. Also, while the fixed camera setting assumed in these methods can suffer from oblique views and limited image resolutions, egocentric setting provides strong appearance cues of people. Our method utilizes ego-motion, scale and pose information to improve the localization performance in such an egocentric setting.


\section{Proposed Method}
\subsection{Overview}
\label{ssec:pf}
\begin{figure*}[t]
  \begin{center}
    \includegraphics[width=\linewidth]{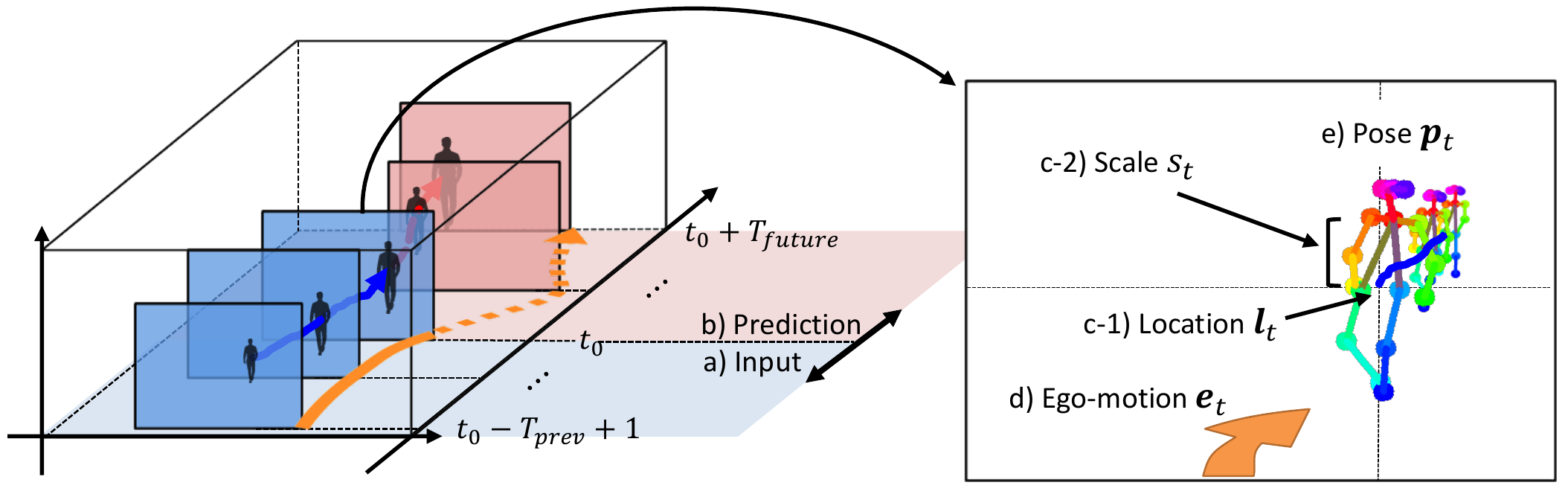}
    \caption{{\bf Future Person Localization in First-Person Videos}. Given a) $T\sub{prev}$-frames observations as input, we b) predict future locations of a target person in the subsequent $T\sub{future}$ frames. Our approach makes use of c-1) locations and c-2) scales of target persons, d) ego-motion of camera wearers and e) poses of the target persons as a salient cue for the prediction.}
    \label{fig:overview}
  \end{center}
\end{figure*}

In this section, we first formulate the problem of predicting future locations of people in first-person videos. Consider a certain \emph{target} person seen in a current frame of a first-person video recorded on the street. Our goal is to predict where the target person will be seen in subsequent frames of the video based on the observation up to the current frame. Formally, let $\vct{l}_{t}\in\mathbb{R}^2_+$ be the 2D location of the person in the frame $t$. As illustrated in Figure~\ref{fig:overview}, we aim to predict the person's relative locations in the subsequent $T\sub{future}$ frames from the current one at $t_0$ (red frames in the figure), that is, $L\sub{out}=(\vct{l}_{t_0+1} - \vct{l}_{t_0}, \vct{l}_{t_0+2} - \vct{l}_{t_0}, ... , \vct{l}_{t_0+T\sub{future}} - \vct{l}_{t_0})$, based on observations in the previous $T\sub{prev}$ frames (blue ones).

\begin{figure}[t]
  \begin{center}
    \includegraphics[width=\linewidth]{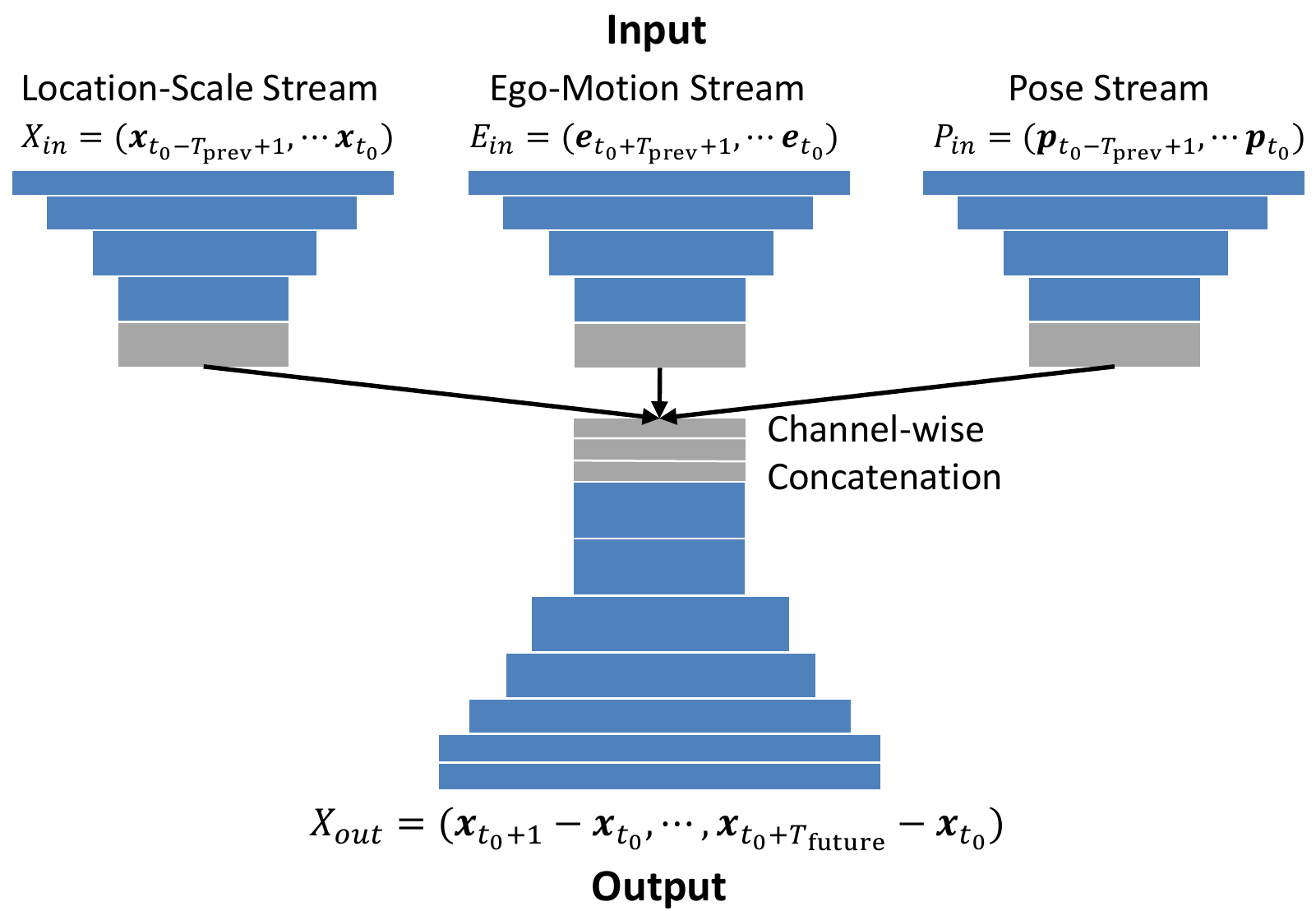}
    \caption{{\bf Proposed Network Architecture}. Blue blocks correspond to convolution/deconvolution layers while gray blocks describe intermediate deep features.}
    \label{fig:model}
  \end{center}
\end{figure}

The key technical interest here is what kind of observations can be used as a salient cue to better predict $L\sub{out}$. Based on the discussions we made in Section~\ref{sec:introduction} (also refer to Figure~\ref{fig:overview}), we focus on c-1) locations and c-2) scales of target people, d) ego-motion of the camera wearer, and e) poses of target people as the cues to approach the problem. In order to predict future locations from those cues, we develop a deep neural network that utilizes a multi-stream convolution-deconvolution architecture shown in Figure~\ref{fig:model}. Input streams take the form of fully-convolutional networks with 1-D convolution filters to learn sequences of the cues shown above. Given a concatenation of features provided from all input streams, the output stream deconvolutes it to generate $L\sub{out}$. The overall network can be trained end-to-end via back-propagation. In the following sections, we describe how each cue is extracted to improve prediction performance. Concrete implementation details and training strategies are discussed in Section~\ref{subsec:implementations}. 

\subsection{Location-Scale Cue}
\label{subsec:pos}
The most straightforward cue to predict future locations of people $L\sub{out}$ is their previous locations up to the current frame $t_0$. For example, if a target person is walking in a certain direction at a constant speed, our best guess based on only previous locations would be to expect them to keep going in that direction in subsequent future frames too. However, visual distances in first-person videos can correspond to different physical distances depending on where people are observed in the frame. 

In order to take into account this perspective effect, we propose to learn both locations and scales of target people jointly. Given a simple assumption that heights of people do not differ too much, scales of observed people can make a rough estimate of how large movements they made in the actual physical world. Formally, let $L\sub{in} = (\vct{l}_{t_0 - T\sub{prev}+1},\dots,\vct{l}_{t_0})$ be a history of previous target locations. Then, we extend each location $\vct{l}_t\in\mathbb{R}^2_+$ of a target person by adding the scale information of that person $s_t\in\mathbb{R}_+$, \ie, $\vct{x}_t=(\vct{l}_t^\top, s_t)^\top$. Then, the \emph{`location-scale'} input stream in Figure~\ref{fig:model} learns time evolution in $X\sub{in}=(\vct{x}_{t_0-T\sub{prev}+1},\dots,\vct{x}_{t_0})$, and the output stream generates $X\sub{out}=(\vct{x}_{{t_0}+1} - \vct{x}_{t_0},\dots,\vct{x}_{{t_0}+T\sub{future}} - \vct{x}_{t_0})$.

\subsection{Ego-Motion Cue}
\label{subsec:ego}
While $X\sub{in}$ explicitly describes how a target person is likely to move over time, the direct prediction of $X\sub{out}$ from $X\sub{in}$ is still challenging due to significant ego-motion present in first-person videos. More specifically, the coordinate system to describe each point $\vct{l}_t$ changes dynamically as the camera wearer moves. This makes $X\sub{in}$ and $X\sub{out}$ quite diverse depending on both walking trajectories of the target person and ego-motion of camera wearers.

Moreover, ego-motion of camera wearers could affect how the target people move as a result of interactive dynamics among people. For instance, consider a case where a target person is walking towards the camera wearer. When the target person and the camera wearer notice that they are going to collide soon, they will explicitly or implicitly condition themselves to change their walking speed and direction to avoid the potential collision. Although some recent work has tried to incorporate such interactive behaviors into human trajectory prediction~\cite{alahi2016social,lee2017desire,Ma_2017_CVPR,Robicquet2016}, their approaches need all interacting people to be observed in a static camera view and cannot be applied directly to our case.

In order to improve future localization performance for first-person videos, we propose to learn how the camera wearer has been moving, \ie, the ego-motion cue. Specifically, we first estimate the rotation and translation between successive frames. Rotation is described by a rotation matrix $R_t\in \mathbb{R}^{3\times 3}$ and translation is described by a 3D vector $\vct{v}_t\in\mathbb{R}^3$ (\ie, x-, y-, z-axes), both from frame $t-1$ to frame $t$ in the camera coordinate system at frame $t-1$. These vectors represent the local movement between the successive frames, however, does not capture the global movement along multiple frames. Therefore, for each frame $t$ within the input interval $[t_0 - T\sub{prev} + 1, t_0]$, we accumulate those vectors to describe time-varying ego-motion patterns in the camera coordinate system at frame $t_0 - T\sub{prev}$:
\begin{eqnarray}
  R'_{t} &=&
  \begin{cases}
    R_{t_0-T\sub{prev}+1} & (t = {t_0}-T\sub{prev}+1) \\
    R_{t-1}~R'_{t} & (t > {t_0}-T\sub{prev}+1),
  \end{cases} \\
  \vct{v}'_t &=& 
  \begin{cases}
    \vct{v}_t & (t = {t_0}-T\sub{prev}+1)\\
    R'^{-1}_t \vct{v}_t + \vct{v}'_{t-1} & (t > {t_0}-T\sub{prev}+1).
  \end{cases}
\end{eqnarray}

We form the feature vector for each frame by concatenating the rotation vector $\vct{r}'_t$ (\ie, yaw, roll, pitch) converted from $R'_t$ and $\vct{v}'_t$, resulting in a 6-dimensional vector $\vct{e}_t$. Finally, we stack them to form an input sequence $E_{in}$ for the \emph{`ego-motion'} stream shown in Figure \ref{fig:model}.

\begin{eqnarray}
    \vct{e}_t &=& ((\vct{r}'_t)^\top, (\vct{v}'_t)^\top)^\top \in \mathbb{R}^6, \\
     E\sub{in} &=& (\vct{e}_{t_0-T\sub{prev}+1},\dots,\vct{e}_{t_0}).
\end{eqnarray}

\subsection{Pose Cue}
\label{subsec:pose}
Another notable advantage of using first-person videos is the ability to observe people up-close. This makes it easier to capture what poses they take (\eg, which directions they orient), which could act as another strong indicator of the direction they are going to walk along.

The \emph{`pose'} stream in Figure~\ref{fig:model} is aimed at encoding such pose information of target people. More specifically, we track temporal changes of several body parts of target people including eyes, shoulders, and hips as a feature of target poses. This results in an input sequence $P\sub{in} = (\vct{p}_{{t_0}-T\sub{prev}+1},\dots,\vct{p}_{t_0})$ where $\vct{p}\in\mathbb{R}_+^{2V}$ is a $2V$-dimensional vector stacking locations of $V$ body parts.

\section{Experiments}
To investigate the effectiveness of our approach in detail, we first construct a new first-person video dataset recorded by a person walking on the street. We also evaluate our method on First-Person Social Interaction Dataset~\cite{fathi2012social} to see if our approach can be applied to a more general case where camera wearers take a variety of actions while walking.

\subsection{First-Person Locomotion Dataset}
\label{ssec:dataset}
\begin{figure}[t]
  \begin{center}
    \includegraphics[width=\linewidth]{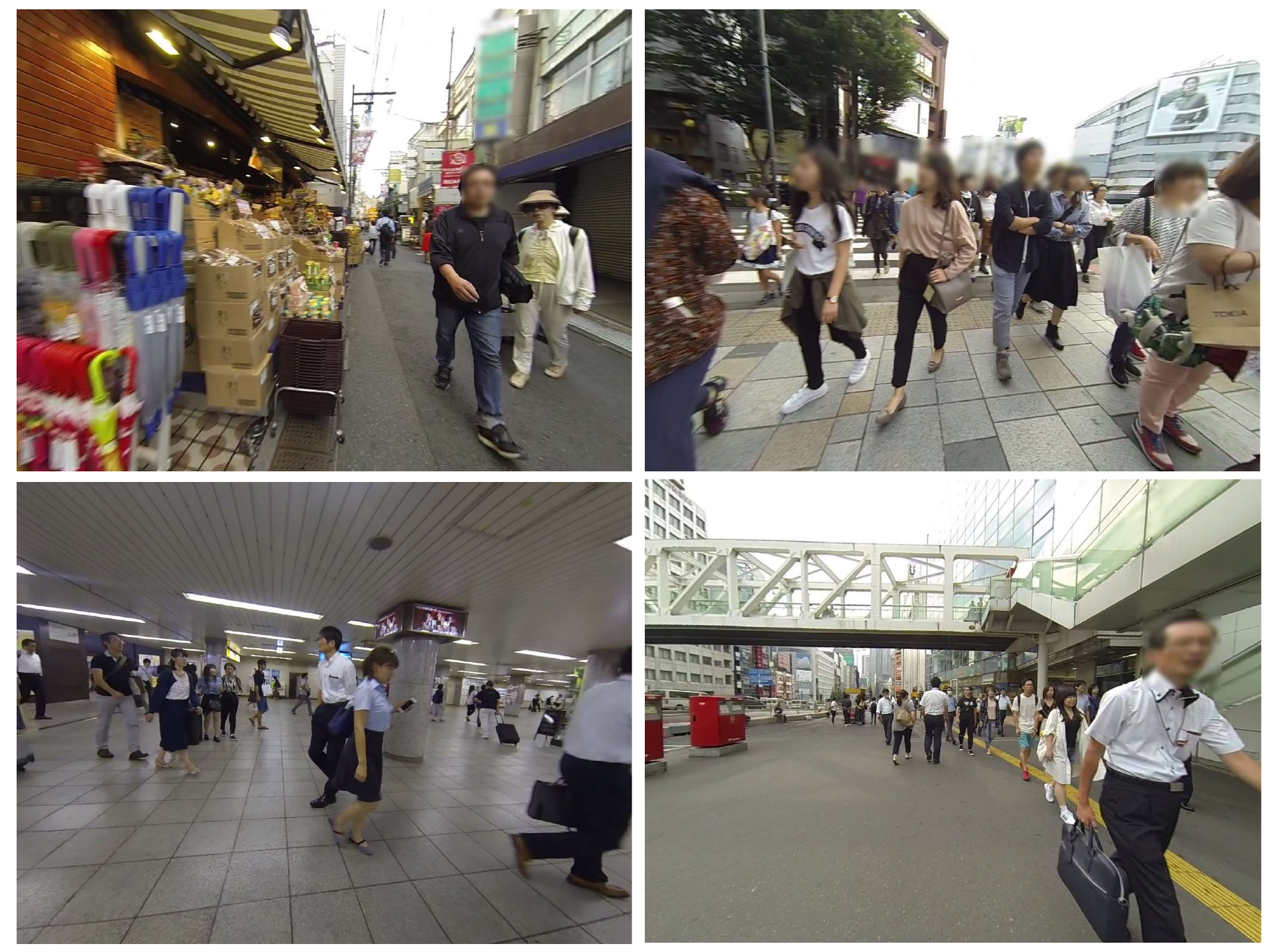}
    \caption{{\bf First-Person Locomotion Dataset} recorded by wearable chest-mounted cameras under diverse environments, which comprises more than 5,000 people in total.}
    \label{fig:dataset}
  \end{center}
\end{figure}

To the best of our knowledge, most of the first-person video datasets comprise scenes where only a limited number of people are observed, \eg, CMU Social Interaction Dataset~\cite{Park2012}, JPL Interaction Dataset~\cite{Ryoo_2013_CVPR}, HUJI EgoSeg Dataset~\cite{Poleg2014}. In this work, we introduce a new dataset which we call \emph{First-Person Locomotion (FPL) Dataset}. The FPL Dataset consists of about 4.5 hours of first-person videos recorded by people wearing a chest-mounted camera and walking around in diverse environments. Some example frames are shown in Figure~\ref{fig:dataset}. The number of observed people is more than 5,000 in total. 

Training and testing samples are given in the form of a tuple $(X\sub{in}, E\sub{in}, P\sub{in}, X\sub{out})$, where $X\sub{in}$ is location-scale, $E\sub{in}$ is camera ego-motion, $P\sub{in}$ is pose, and $X\sub{out}$ is relative future location-scale with respect to $\vct{x}_{t_0}$. $X\sub{in},E\sub{in}, P\sub{in}$ are available both in training and testing times and defined in interval $[{t_0}- T\sub{prev} + 1, {t_0}]$. On the other hand, $X\sub{out}$ serves as ground-truth defined in $[{t_0} + 1,\dots, t_0 + T\sub{future}]$, which we can access only during the training time. In this experiment, we set $T\sub{prev} = T\sub{future} = 10$ at 10 fps, \ie, a time window of one second for both observation and prediction.

We generated the samples as follows. For each frame, we detected people with OpenPose~\cite{cao2017realtime}. We tracked the upper body of detected people over time using the kernelized correlation filter~\cite{henriques2015high} after two consecutive frames were aligned with homography. We terminated the tracking if subsequent detection results were not found within a certain pre-defined spatiotemporal range. As a result of this tracking, we obtained many short tracklets\footnote{Out of 830,000 human poses detected first, approximately 200,000 (24.1\%) poses were successfully associated to form the valid samples.}. These tracklets were then merged to generate longer ones with the conditions 1) if the detected person at the tail of one tracklet is visually similar to that at the head of the other tracklet and 2) if these tracklets were also spatiotemporally close enough. A cosine distance of deep features extracted by Faster R-CNN~\cite{renNIPS15fasterrcnn} was used to measure visual similarity.

For each tracklet, we extracted locations $\vct{l}_t$, scales $s_t$, poses $\vct{p}_t$, and ego-motion $\vct{e}_t$ as follows. First, we extracted 18 body parts using OpenPose~\cite{cao2017realtime}. $\vct{l}_t$ was then defined by the middle of two hips. Also, $s_t$ was given by the distance between the location of the neck and $\vct{l}_t$. Furthermore, we obtained $\vct{p}_t$ as a $36$-dimensional feature (\ie, $V=18$), which was normalized by subtracting $\vct{l}_t$ and divided by $s_t$. $\vct{e}_t$ was estimated by the unsupervised ego-motion estimator~\cite{zhou2017unsupervised}. Finally, we applied sliding window to generate multiple fixed length (\ie, 2 seconds) samples. As a result of this procedure, we obtained approximately 50,000 samples in total. 

\subsection{Implementation Details}
\label{subsec:implementations}

\paragraph{Architecture choice}
The full specification of the proposed network architecture is shown in Table~\ref{tab:model}. Each input stream feeds $D\times 10$-dimensional inputs (where $D$ changes depending on which cues we focus on) to four cascading 1D temporal convolution layers of different numbers of channels, each of which is followed by batch normalization (BN)~\cite{ioffe2015batch} and rectified linear unit (ReLU) activation~\cite{icml2010_NairH10}. Then, $128\times 2$-dimensional features from the input streams are concatenated and fed to the output stream consisting of two 1D convolution layers with BN and ReLU, four cascading 1D deconvolution layers also with BN and ReLU, and one another 1D convolution layer with linear activation.

\begin{table}[t]
\begin{center}
\scalebox{0.9}{
\begin{tabular}{@{}lrrr@{}}
\toprule
Layer type & Channel & Kernel size & Output size \\
\midrule
\multicolumn{4}{c}{Input streams (Location-scale, ego-motion, and pose)} \\
\midrule
Input & - & - & $D\times10$ \\
1D-Conv+BN+ReLU & 32 & 3 & $32\times8$ \\
1D-Conv+BN+ReLU & 64 & 3 & $64\times6$ \\
1D-Conv+BN+ReLU & 128 & 3 & $128\times4$ \\
1D-Conv+BN+ReLU & 128 & 3 & $128\times2$ \\
\midrule
\multicolumn{4}{c}{Output stream} \\
\midrule
Concat & - & - & $384\times2$ \\
1D-Conv+BN+ReLU & 256 & 1 & $256\times2$ \\
1D-Conv+BN+ReLU & 256 & 1 & $256\times2$ \\
1D-Deconv+BN+ReLU & 256 & 3 & $256\times4$ \\
1D-Deconv+BN+ReLU & 128 & 3 & $128\times6$ \\
1D-Deconv+BN+ReLU & 64 & 3 & $64\times8$ \\
1D-Deconv+BN+ReLU & 32 & 3 & $32\times10$ \\
1D-Conv+Linear & 3 & 1 & $3\times10$ \\
\bottomrule \\
\end{tabular}
}
\end{center}
\caption{{\bf Our Network Architecture} where BN: batch normalization~\cite{ioffe2015batch} and ReLU: rectifier linear unit~\cite{icml2010_NairH10}. The network consists of three input streams and one output stream, where inputs have different dimensions $D$ depending on the streams: $D=3$ for the location-scale stream, $D=6$ for the ego-motion stream, and $D=36$ for the pose stream. }
\label{tab:model}
\end{table}

\paragraph{Optimization}
To train the network, we first normalized $X\sub{in}$ and $X\sub{out}$ to have zero-mean and unit variance. We also adopted a data augmentation by randomly flipping samples horizontally. The loss functions to predict $X\sub{out}$ was defined by the mean squared error (MSE). We optimized the network via Adam~\cite{Diederik2014} for 17,000 iterations with mini-batches of 64 samples, where a learning rate was initially set to 0.001 and halved at 5,000, 10,000, 15,000 iterations. All implementations were done with Chainer~\cite{chainer_learningsys2015}. 

\subsection{Evaluation Protocols}
\label{subsec:eval_protocol}
\paragraph{Data splits}
We adopted five-fold cross-validation by randomly splitting samples into five subsets. We ensured that samples in training and testing subsets were drawn from different videos. Training each split required about 1.5 hours on a single NVIDIA TITAN X. Also when evaluating methods with testing subsets, we further divided samples into three conditions based on how people walked (\ie, walking directions): target people walked a) {\bf Toward}, b) {\bf Away} from, or c) {\bf Across} the view of a camera. Further details on how to segregate the samples into these three categories are present in our supplementary materials.

\paragraph{Evaluation metric}
Although our network predicts both locations and scales of people in the future frames, we measured its performance based on how accurate the predicted locations were. Similar to \cite{alahi2016social}, we employed the final displacement error (FDE) as our evaluation metric. Specifically, FDE was defined by the L2 distance between predicted final locations $\vct{l}_{t_0 + T\sub{future}}$ and the corresponding ground-truth locations.

\paragraph{Baselines}
Since there were no prior methods that aimed to predict future person locations in first-person videos, we have implemented the following baselines.
\begin{itemize}
\item {\bf ConstVel}: Inspired by the baseline used in \cite{ParkCVPR16}, this method assumes that target people moved straight at a constant speed. Specifically, we computed the average speed and direction from $X\sub{in}$ to predict where the target would be located at the ${t_0}+T\sub{future}$-th frame.
\item {\bf NNeighbor}: We selected $k$-nearest neighbor input sequences in terms of the L2 distance on the sequences of locations $L\sub{in}$ and derived the average of $k$-corresponding locations at frame ${t_0}+T\sub{future}$. In our experiments, we set $k=16$ as it performed well.
\item {\bf Social LSTM~\cite{alahi2016social}}: We also evaluated Social LSTM, one of the state-of-the-art approaches on human trajectory prediction, with several minor modifications to better work on first-person videos. Specifically, we added the scale information to inputs and outputs. The estimation of Gaussian distributions was replaced by direct prediction of $X_{out}$ as it often failed on the FPL Dataset. The neighborhood size $N_o$ used in the paper was set to $N_o=256$. 
\end{itemize}

\subsection{Results}

\paragraph{Quantitative evaluation}
Table~\ref{tab:analysis} reports FDE scores on our FPL Dataset. Overall, all methods were able to predict future locations of people with the FDE less than about 15\% of the frame width (approximately $19^\circ$ in horizontal angle). We confirmed that our method ({\bf Ours}) has significantly outperformed the other baselines. Since walking speeds and directions of people were quite diverse and changing dynamically over time, naive baselines like {\bf ConstVel} and {\bf NNeighbor} did not perform well. Moreover, we found that {\bf Social LSTM}~\cite{alahi2016social} performed poorly. Without explicitly taking into account how significant ego-motion affects people locations in frames, temporal models like LSTM would not be able to learn meaningful temporal dynamics, ultimately rendering their predictions quite unstable. Note that without our modification shown in Section~\ref{subsec:eval_protocol}, the performance of vanilla Social LSTM was further degraded (\ie, 152.87 FDE on average). Comparing results among walking directions, {\bf Toward} was typically more challenging than other conditions. This is because when target people walked toward the view of a camera, they would appear in the lower part of frames, making variability of future locations much higher than other walking directions.

\paragraph{Error analysis} We investigated the distribution of the errors. With our method, 73\% samples received error less than 100 pixels ($10^\circ$ in horizontal angle). There were only 1.4\% samples suffered from significant error larger than 300 pixels ($30^\circ$ in horizontal angle). Additionally, we calculated the errors normalized by each sample's scale. By assuming that the length between the center hip and the neck of a person to be 60\,cm, the average error obtained by our method approximately corresponded to 60\,cm in the physical world.

\begin{table}[t]
\centering
\begin{tabular}{@{}lcccc@{}}
\toprule
Method & \multicolumn{4}{c}{\raisebox{.5mm}{Walking direction}} \\ \cline{2-5}
& \raisebox{-.5mm}{Toward} & \raisebox{-.5mm}{Away} & \raisebox{-.5mm}{Across} & \raisebox{-.5mm}{Average} \\
\midrule
ConstVel & 178.96 & 98.54 & 121.60 & 107.15 \\
NNeighbor  & 165.78 & 89.81 & 123.83 & 98.38 \\
Social LSTM\cite{alahi2016social} & 173.02 & 111.24 & 148.83 & 118.10 \\
\midrule
{\bf Ours} & {\bf 109.03} & {\bf 75.56} & {\bf 93.10} & {\bf 77.26} \\
\bottomrule\\
\end{tabular}
\caption{{\bf Comparisons to Baselines}. Each score describes the final displacement error (FDE) in pixels with respect to the frame size of $1280\times 960$-pixels.}
\label{tab:analysis}
\end{table}

\begin{table}[t]
\centering
\scalebox{.92}{
\begin{tabular}{@{}lcccc@{}}
\toprule
Method & \multicolumn{4}{c}{\raisebox{.5mm}{Walking direction}} \\ \cline{2-5}
& \raisebox{-.5mm}{Toward} & \raisebox{-.5mm}{Away} & \raisebox{-.5mm}{Across} & \raisebox{-.5mm}{Average} \\
\midrule
$L\sub{in}$ & 147.23 & 80.90 & 104.85 & 88.16 \\
$X\sub{in}$ & 126.64 & 79.09 & 102.98 & 81.86 \\
$X\sub{in} + E\sub{in}$ & 122.16 & 76.67 & 99.39 & 79.09 \\
$X\sub{in} + P\sub{in}$ & 113.33 & 78.55 & 100.33 & 80.57 \\
\midrule
{\bf Ours} \small{($X\sub{in} + E\sub{in} + P\sub{in}$)}& {\bf 109.03} & {\bf 75.56} & {\bf 93.10} & {\bf 77.26} \\
\bottomrule\\
\end{tabular}
}
\caption{{\bf Ablation Study}. $L\sub{in}$: locations, $X\sub{in}$ location-scales, $E\sub{in}$: ego-motion, and $P\sub{in}$: poses. Each score describes the final displacement error (FDE) in pixels with respect to the frame size of $1280\times 960$-pixels.}
\label{tab:ablation}
\end{table}
 
\paragraph{Qualitative evaluation}
Figure~\ref{fig:qual} presents several visual examples of how each method worked. Examples (a), (b), and (c) are results drawn respectively from {\bf Toward}, {\bf Across}, and {\bf Away} subsets. Especially, significant ego-motion of the camera wearer to turn right was observed in Example (b), making predictions of baseline methods completely failure. Another case where ego-motion played an important role was when target people did not move, such as the person standing still in Example (d). Example (e) involves not only significant ego-motion but also changes in walking direction of the target. Our method successfully performed in this case as it could capture postural changes of target persons for prediction.

\begin{figure*}[t]
\centering
    \includegraphics[width=.95\linewidth]{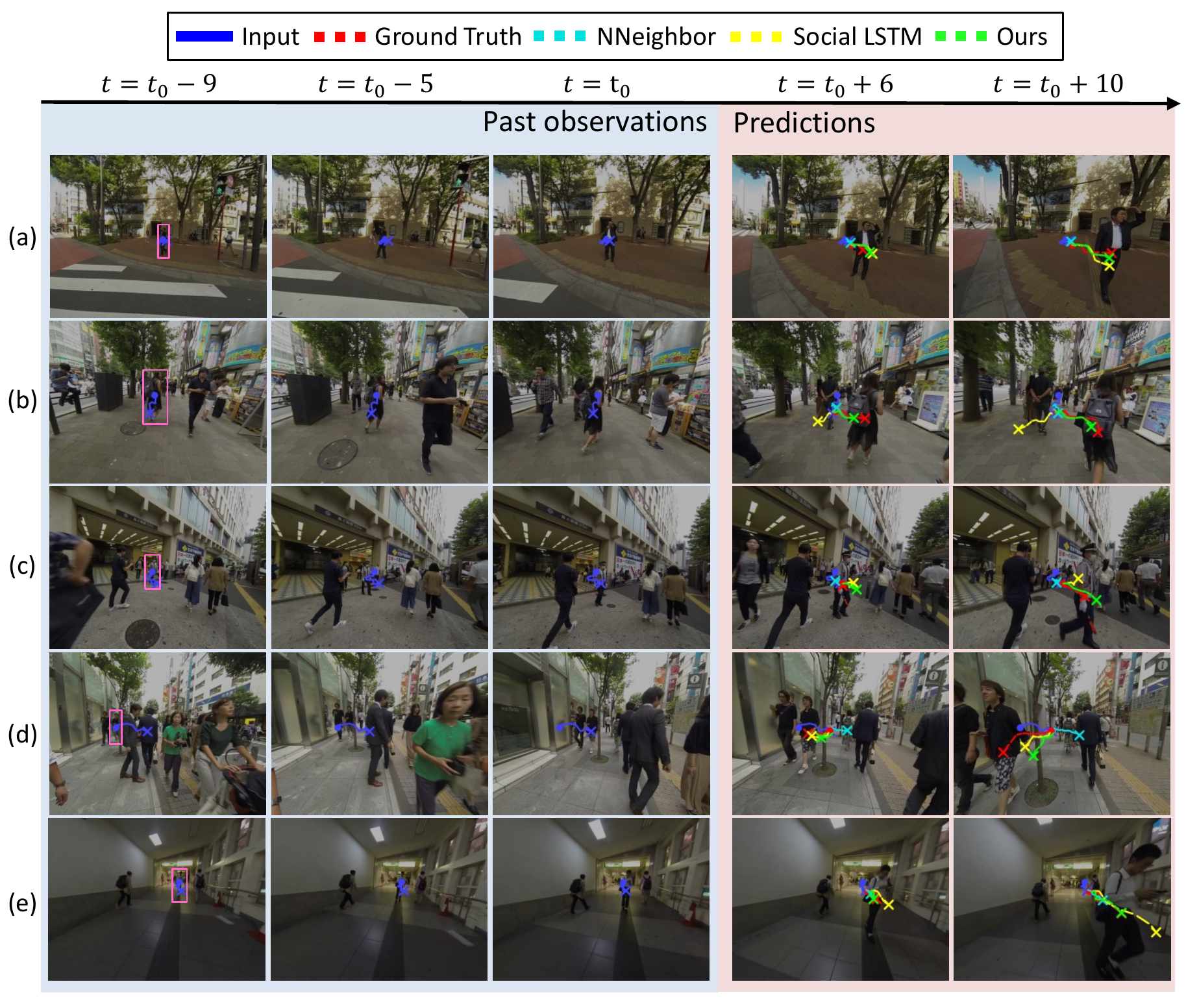}
    \caption{{\bf Visual Examples of Future Person Localization}. Using locations (shown with solid blue lines), scales and poses of target people (highlighted in pink, left column) as well as ego-motion of camera wearers in the past observations highlighted in blue, we predict locations of that target (the ground-truth shown with red crosses with dotted red lines) in the future frames highlighted in red. We compared several methods: {\bf Ours} (green), {\bf NNeighbor} (cyan), and {\bf Social LSTM}~\cite{alahi2016social} (yellow).}
\label{fig:qual}
\end{figure*}

\paragraph{Ablation study}
We made an ablation study to see how each of scales, ego-motion, and poses contributed overall prediction performances. Specifically, we started from the only location information $L\sub{in}$, then added scale information to use $X\sub{in}$. For these two conditions, we learned a single-stream convolution-deconvolution architecture. Then, we evaluated the combination of $X\sub{in} + E\sub{in}$ (locations, scales, and ego-motion) and that of $X\sub{in} + P\sub{in}$ (locations, scales, and poses) by learning two-stream convolution-deconvolution architectures. Results are shown in Table~\ref{tab:ablation}. We confirmed that all of the cues helped individually to improve prediction performances. Especially, significant performance gains were observed on the {\bf Toward} subset from $L\sub{in}$ to $X\sub{in}$, \ie, by introducing scale information, and from $X\sub{in}$ to $X\sub{in}+P\sub{in}$, \ie, by further combining pose information.

\begin{figure}[t]
\begin{center}
\end{center}
    \includegraphics[width=\linewidth]{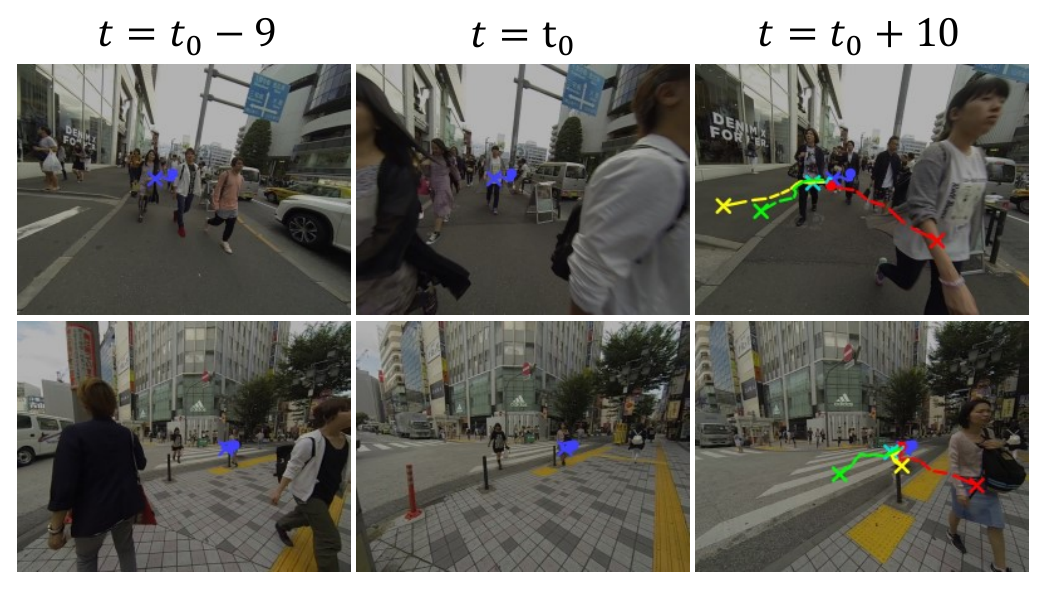}
    \caption{{\bf Failure Cases}. Given previous locations (blue), predictions by our method (green) and Social LSTM~\cite{alahi2016social} (yellow) both deviated from ground-truth future locations (red).}
\label{fig:failed}
\end{figure}

\paragraph{Failure cases and possible extensions}
Figure~\ref{fig:failed} shows several typical failure cases. On both examples, our method and other baselines did not perform accurately as camera wearers made sudden unexpected ego-motion. One possible solution to cope with these challenging cases is to predict future movements of the camera wearers as done in \cite{ParkCVPR16}.

\subsection{Evaluation on Social Interaction Dataset}
\label{ssec:si}

Finally, we evaluate how our approach works on First-Person Social Interaction Dataset~\cite{fathi2012social}. This dataset consists of several first-person videos taken in an amusement park and involves a variety of social moments like communicating with friends, interacting with a clerk, and waiting in line, standing for a more general and challenging dataset. In our experiment, we manually extracted a subset of videos where camera wearers kept walking while sometimes interacting with others. From this subset, we collected approximately 10,000 samples in total. Similar to the previous experiment, we adopted five-fold cross-validation to evaluate how our method and other baselines performed.

\paragraph{Training setup}
In this dataset, camera wearers frequently turned their head to pay their attention to various different locations. This made ego-motion estimator~\cite{zhou2017unsupervised} completely inaccurate as it was originally trained to estimate ego-motion of vehicle-mounted cameras, where such frequent turning was hardly observed in their training datasets. To cope with this, instead of the velocity and rotation used in Section~\ref{subsec:ego}, we made use of optical flows to describe ego-motion cues. More specifically, we computed dense optical flows using~\cite{ilg2017flownet} and divided them into $4\times 3$ grids. We then computed average flows per grid and concatenate them to obtain 24-dimensional vector for describing ego-motion per frame. For the training, we first pre-trained our network on FPL Dataset with the same training strategies shown in Section~\ref{subsec:implementations} but with the above flow-based ego-motion feature \footnote{Our network with flow-based features resulted in $79.15$ FDE on FPL dataset, \ie, $1.89$ performance drop from the original result shown in Table~\ref{tab:analysis}. One possible reason for the better performance using ego-motion features based on \cite{zhou2017unsupervised} is that they can capture yaw rotations (\ie, turning left and right) of camera wearers more accurately.}. We then fine-tuned this trained network on the Social Interaction Dataset for 200 iterations using Adam with a learning rate of 0.002.

\begin{figure}[t]
\begin{center}
\end{center}
    \includegraphics[width=\linewidth]{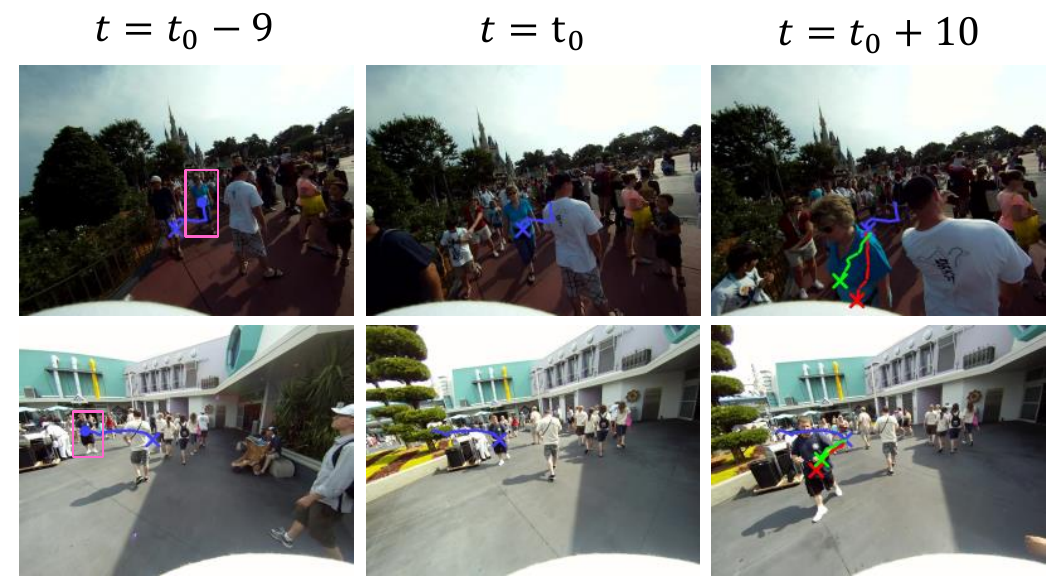}
    \caption{{\bf Visual Examples from Social Interaction Dataset~\cite{fathi2012social}}: previous locations (blue lines) of target people (pink bounding boxes); predictions by our method (green lines); and ground-truth future locations (red lines).}
\label{fig:si_result}
\end{figure}

\paragraph{Results}
FDE scores are shown in Table~\ref{tab:si_result}. Similar to the previous experiment, we divided testing datasets into three subsets, Toward, Away, and Across, based on walking directions of target people. Although performances of all methods were rather limited compared to the previous results in Table~\ref{tab:analysis}, we still confirmed that our method was able to outperform other baseline methods including Social LSTM~\cite{alahi2016social}. Some visual examples are also shown in Figure~\ref{fig:si_result}.

\begin{table}[t]
\centering
\begin{tabular}{@{}lcccc@{}}
\toprule
Method & \multicolumn{4}{c}{\raisebox{.5mm}{Walking direction}} \\ \cline{2-5}
& \raisebox{-.5mm}{Toward} & \raisebox{-.5mm}{Away} & \raisebox{-.5mm}{Across} & \raisebox{-.5mm}{Average} \\
\midrule
ConstVel & 173.75 & 176.76 & 133.32 & 170.71 \\
NNeighbor & 167.11 & 159.26 & 148.91 & 162.02 \\
Social LSTM~\cite{alahi2016social} & 240.03 & 196.48 & 223.37 & 213.59 \\
\midrule
{\bf Ours}& {\bf 131.94} & {\bf 125.48} & {\bf 112.88} & {\bf 125.42} \\
\bottomrule \\ 
\end{tabular}
\caption{{\bf Results on Social Interactions Dataset \cite{fathi2012social}.} Each score describes the final displacement error (FDE) in pixels with respect to the frame sizes of either $1280\times 960$-pixels or $1280\times 720$-pixels.}
\label{tab:si_result}
\end{table}

\section{Conclusion}
We have presented a new task called future person localization in first-person videos. Experimental results have revealed that ego-motion of camera wearers as well as scales and poses of target people were all necessary ingredients to accurately predict where target people would appear in future frames.

As we discussed with the failure cases, one possible direction for extending this work is to incorporate future localization of camera wearers~\cite{ParkCVPR16}. By knowing how the camera wearers move in the near future, we should be able to predict future locations of observed people more accurately in first-person videos. 

\appendix
\section*{Appendix}
\section{Data Statistics}
\begin{figure*}[t]
\begin{center}
    \includegraphics[width=\linewidth]{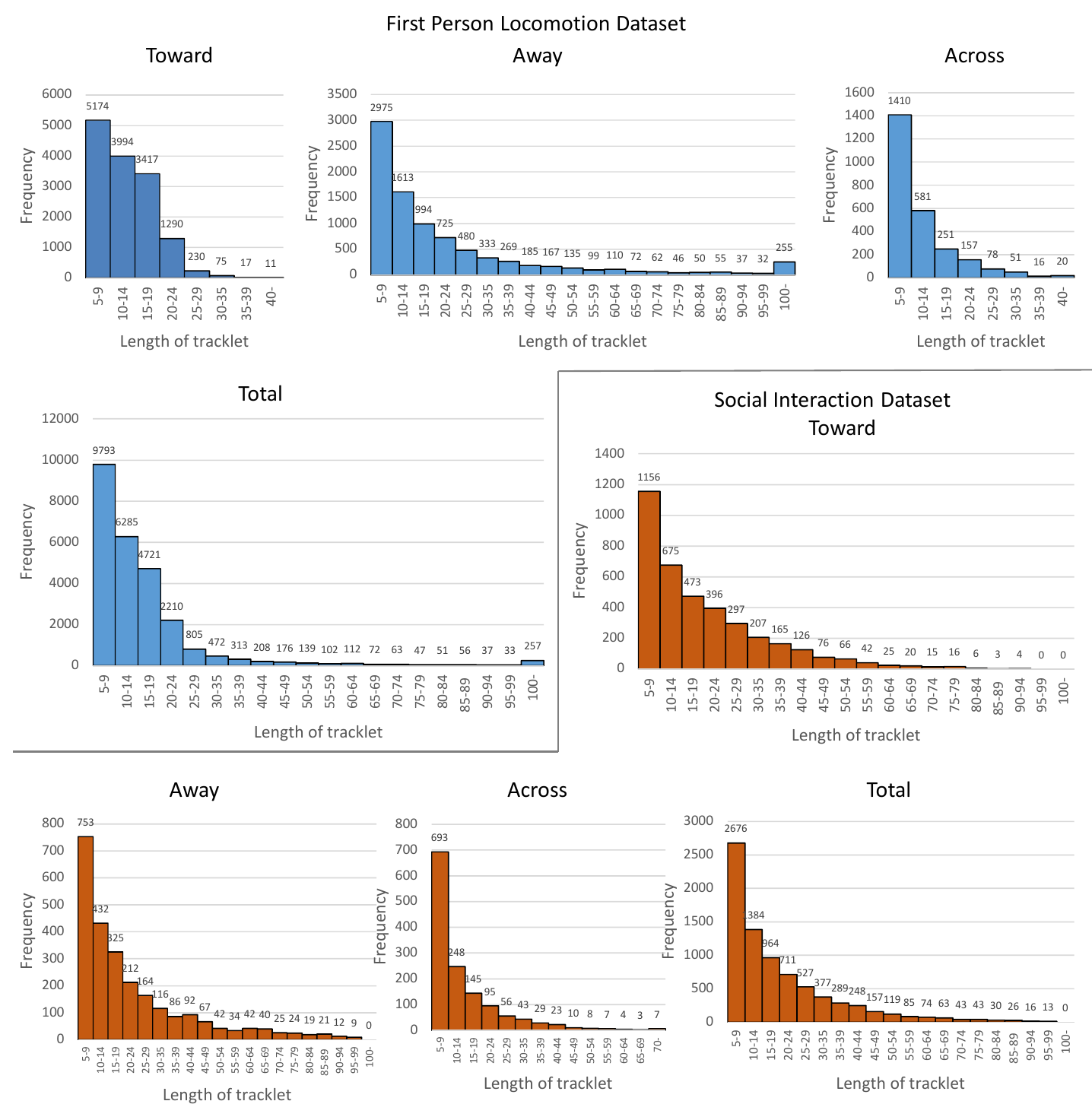}
    \caption{{\bf Distributions of Tracklet Lengths}. Frequency distributions of various lengths of tracklets extracted from First-Person Locomotion Dataset and Social Interaction Dataset \cite{fathi2012social} for three walking directions and the entire database, respectively.}
    \label{fig:stats}
\end{center}
\end{figure*}

Figure~\ref{fig:stats} presents frequency distributions of lengths of the tracklets extracted from First-Person Locomotion Dataset and Social Interaction Dataset~\cite{fathi2012social}. These statistics revealed that most people appeared only for a short time period. In our experiments, we tried to pick out tracklets which were 1) longer enough to learn meaningful temporal dynamics and 2) observed frequently in the datasets to stably learn our network. These requirements resulted in our 50,000 samples consisting of the tracklets longer than or equal to 20 frames (\ie, 2 seconds at 10 fps) and our problem setting of `predicting one-second futures from one-second histories'.

\paragraph{Details of sample division:}We first calculated the mean of scale normalized lengths between the left hip and the right hip for the target person. If this mean is less than 0.25 we categorized the clip as {\bf Across}.
In the remaining clips, we labeled each frame of the clip as either {\bf Toward} if X-coordinate of the left hip is larger than that of the right hip and {\bf Away} otherwise. If the number of frames labeled {\bf Toward} is more than 75\% of the total number of frames in the clip, the clip is categorized as {\bf Toward} and as {\bf Away} if it is less than 25\%.

\begin{table}[t]
\begin{center}
\scalebox{.9}{
\begin{tabular}{@{}llcccc@{}}
\toprule
$T\sub{prev}$ & $T\sub{future}$ & \multicolumn{4}{c}{\raisebox{.5mm}{Walking direction}} \\ \cline{3-6}
& & \raisebox{-.5mm}{Toward} & \raisebox{-.5mm}{Away} & \raisebox{-.5mm}{Across} & \raisebox{-.5mm}{Average} \\
\hline\hline
6 & 10 & 111.39 & 78.54 & 98.41 & 79.77 \\
10 & 10 & 109.03 & 75.56 & 93.10 & 77.26 \\
\hline
6 & 6 & 53.12 & 46.49 & 52.75 & 46.16 \\
10 & 6 & 52.69 & 46.10 & 53.15 & 45.92 \\
\bottomrule
\end{tabular}
}
\end{center}
\caption{{\bf Different Input/Output Lengths}. Final Displacement Error (FDE) for various combinations of input ($T\sub{prev}$) and output ($T\sub{future}$) lengths.}
\label{tab:ablation_time}
\end{table}

\begin{table}[t]
\begin{center}
\scalebox{.9}{
\begin{tabular}{@{}lcccc@{}}
\toprule
$T\sub{prev}$ & \multicolumn{4}{c}{\raisebox{.5mm}{Walking direction}} \\ \cline{2-5}
& \raisebox{0mm}{Toward} & \raisebox{0mm}{Away} & \raisebox{0mm}{Across} & \raisebox{0mm}{Average}\rule[0mm]{0mm}{4mm} \\
\hline\hline
Social LSTM \cite{alahi2016social} & 299.81 & 222.30 & 236.48 & 223.16 \\
\midrule
{\bf Ours} & {\bf 184.62} & {\bf 125.41} & {\bf 169.01} & {\bf 124.85} \\
\bottomrule
\end{tabular}
}
\end{center}
\caption{{\bf Predicting Two-Second Futures}. Final Displacement Error (FDE) where $T\sub{future}$ was set to $T\sub{future}=20$.}
\label{tab:two_sec_prediction}
\end{table}

\begin{table}[t]
\begin{center}
\scalebox{.9}{
\begin{tabular}{@{}lcccc@{}}
\toprule
Method & \multicolumn{4}{c}{\raisebox{.5mm}{Walking direction}} \\ \cline{2-5}
& \raisebox{-.5mm}{Toward} & \raisebox{-.5mm}{Away} & \raisebox{-.5mm}{Across} & \raisebox{-.5mm}{Average} \\
\hline\hline
$X_{in}$ & 136.43 & {\bf 124.10} & 117.56 & 127.40 \\
$X_{in}+E_{in}$ &136.52 & 124.22 & 115.00 & 127.28 \\
$X_{in}+P_{in}$ & 133.10 & 124.57 & 114.80 & 125.78 \\
\midrule
{\bf Ours ($X\sub{in} + E\sub{in} + P\sub{in}$)} & {\bf 131.94} & 125.48 & {\bf 112.88} & {\bf 125.42} \\
\bottomrule
\end{tabular}
}
\end{center}
\caption{{\bf Ablation Study on Social Interactions Dataset \cite{fathi2012social}}. Final displacement error (FDE) for various combination of input features. Notations were the same as those of Table \ref{tab:two_sec_prediction}.}
\label{tab:si_result}
\end{table}

\section{Additional Results}
\subsection{Other Choices of Input/Output Lengths}
In our experiments, we fixed the input and output lengths $T\sub{prev}, T\sub{future}$ to be $T\sub{prev}=T\sub{future}=10$. Table~\ref{tab:ablation_time} shows how performances changed for other choices of $T\sub{prev}$ and $T\sub{future}$. Overall, longer input lengths led to better performance ($T\sub{prev}=6$ vs. $10$). Also, predicting more distant futures becomes more difficult ($T\sub{future}=10$ vs. $6$). To receive shorter inputs, we applied 1-padding to the first and second convolution layer in each stream.

We also compared our method against Social LSTM \cite{alahi2016social} on the task of predicting two-second futures (\ie, $T\sub{future}=20$) in Table~\ref{tab:two_sec_prediction}. We confirmed that our method still worked well on this challenging condition. To generate 20 frame prediction, we changed the kernel size of the deconvolution layers of 3, 3, 3, 3 to 3, 5, 7, 7.

\subsection{Other Visual Examples}
Figure \ref{fig:qual} shows additional visual examples of how our method, as well as several baselines, predicted future locations of people.

\begin{figure*}[t]
\centering
    \includegraphics[width=\linewidth]{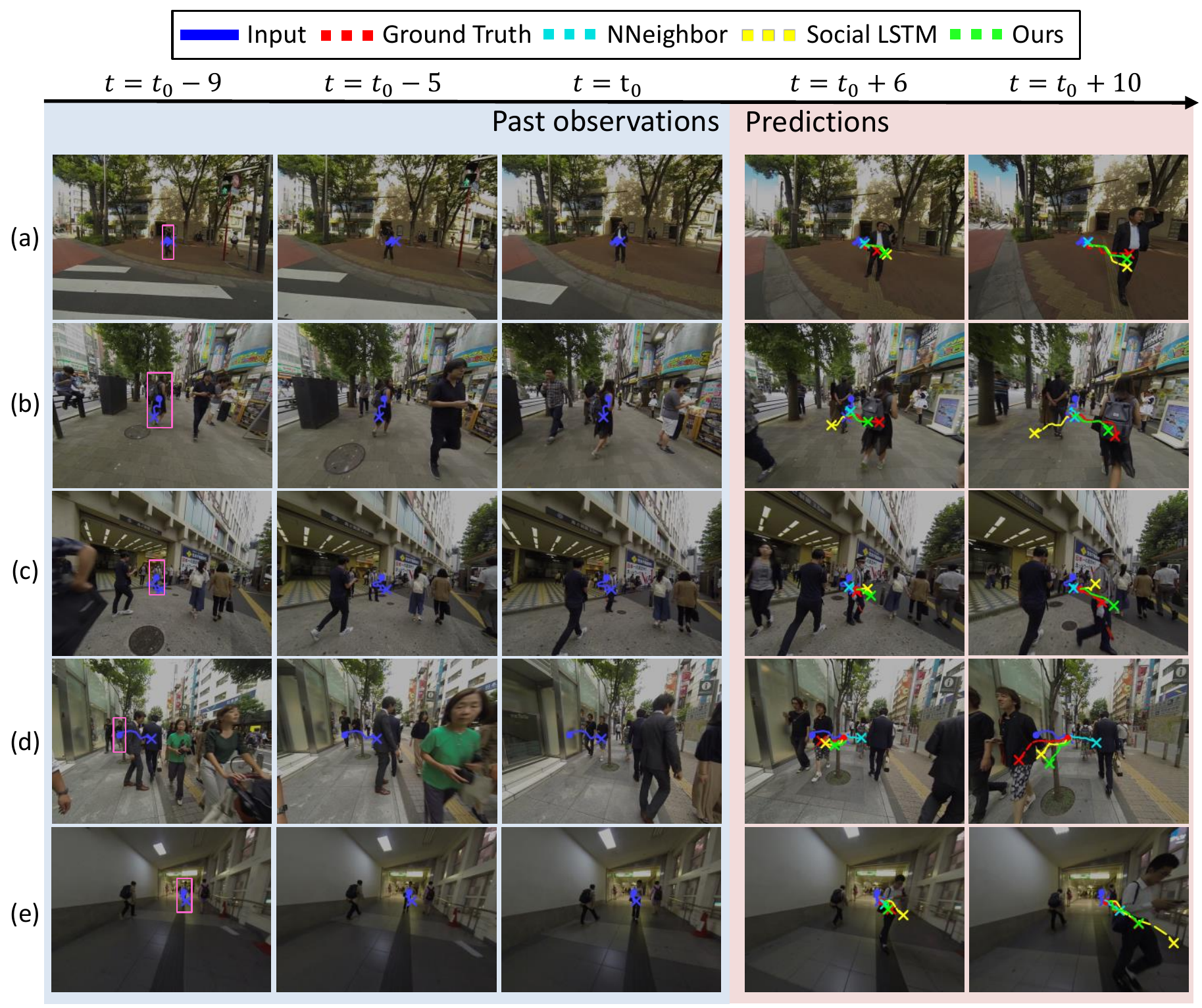}
    \caption{{\bf Qualitative Examples of Future Person Localization on First Person Locomotion Dataset}. (Row 1) Even though input sequence is almost static, our model is able to capture the left turn caused by the wearer's ego-motion. (Row 2, 3) In the input sequence, the target is changing the pose to move right. While compared model fails to predict because of being agnostic to the pose information, our model produces a better prediction. (Row 4) The behavior with respect to complicated ego-motion. In the input sequence, the wearer is turning left to avoid other pedestrians. However, in the future frames, the wearer moves to the opposite side to avoid contact with the target. In this case, our prediction is perturbed due to ego-motion and predicts worse than Social LSTM. (Row 5) Our model works well both in outdoor scenes as well as indoor scenes.}
\label{fig:qual}
\end{figure*}

\subsection{Ablation Study on Social Interaction Dataset}
We performed an ablation study on Social Interaction Dataset~\cite{fathi2012social} in Table~\ref{tab:si_result}. While we computed ego-motion based on optical flows, the combination of ego-motion and pose cues contributed to performance improvements.

\section*{Acknowledgments}
This work was supported by JST CREST Grant Number JPMJCR14E1, Japan.

\clearpage


\balance
{\small
\bibliographystyle{ieee}

\begin{thebibliography}{10}\itemsep=-1pt

\bibitem{alahi2016social}
A.~Alahi, K.~Goel, V.~Ramanathan, A.~Robicquet, L.~Fei-Fei, and S.~Savarese.
\newblock Social lstm: Human trajectory prediction in crowded spaces.
\newblock In {\em Proceedings of the IEEE Conference on Computer Vision and
  Pattern Recognition}, pages 961--971, 2016.

\bibitem{Alletto2015}
S.~Alletto, G.~Serra, S.~Calderara, and R.~Cucchiara.
\newblock Understanding social relationships in egocentric vision.
\newblock {\em Pattern Recognition}, 48(12):4082--4096, 2015.

\bibitem{Bambach2015}
S.~Bambach, S.~Lee, D.~J. Crandall, and C.~Yu.
\newblock Lending a hand: Detecting hands and recognizing activities in complex
  egocentric interactions.
\newblock In {\em Proceedings of the IEEE International Conference on Computer
  Vision}, pages 1949-- 1957, 2015.

\bibitem{Cai2015}
M.~Cai, K.~M. Kitani, and Y.~Sato.
\newblock A scalable approach for understanding the visual structures of hand
  grasps.
\newblock In {\em Proceedings of the IEEE International Conference on Robotics
  and Automation}, pages 1360--1366, 2015.

\bibitem{cao2017realtime}
Z.~Cao, T.~Simon, S.-E. Wei, and Y.~Sheikh.
\newblock Realtime multi-person 2d pose estimation using part affinity fields.
\newblock In {\em Proceedings of the IEEE Conference on Computer Vision and
  Pattern Recognition}, pages 7291 -- 7299, 2017.

\bibitem{fan2017forecasting}
C.~Fan, J.~Lee, and M.~S. Ryoo.
\newblock Forecasting hand and object locations in future frames.
\newblock {\em CoRR}, abs/1705.07328, 2017.

\bibitem{Fathi2011}
A.~Fathi, A.~Farhadi, and J.~M. Rehg.
\newblock {Understanding Egocentric Activities}.
\newblock In {\em Proceedings of the IEEE InternationalConference on Computer
  Vision}, pages 407--414, 2011.

\bibitem{fathi2012social}
A.~Fathi, J.~K. Hodgins, and J.~M. Rehg.
\newblock Social interactions: A first-person perspective.
\newblock In {\em Proceedings of the IEEE Conference on Computer Vision and
  Pattern Recognition}, pages 1226--1233, 2012.

\bibitem{Furnari:2017:NPE:3163595.3163822}
A.~Furnari, S.~Battiato, K.~Grauman, and G.~M. Farinella.
\newblock Next-active-object prediction from egocentric videos.
\newblock {\em Journal of Visual Communication and Image Representation},
  49(C):401--411, 2017.

\bibitem{henriques2015high}
J.~F. Henriques, R.~Caseiro, P.~Martins, and J.~Batista.
\newblock High-speed tracking with kernelized correlation filters.
\newblock {\em IEEE Transactions on Pattern Analysis and Machine Intelligence},
  37(3):583--596, 2015.

\bibitem{Hoshen2016}
Y.~Hoshen and S.~Peleg.
\newblock An egocentric look at video photographer identity.
\newblock In {\em Proceedings of the IEEE Conference on Computer Vision and
  Pattern Recognition}, pages 4284--4292, 2016.

\bibitem{Huang2016}
S.~Huang, X.~Li, Z.~Zhang, Z.~He, F.~Wu, W.~Liu, J.~Tang, and Y.~Zhuang.
\newblock Deep learning driven visual path prediction from a single image.
\newblock {\em IEEE Transactions on Image Processing}, 25(12):5892--5904, 2016.

\bibitem{ilg2017flownet}
E.~Ilg, N.~Mayer, T.~Saikia, M.~Keuper, A.~Dosovitskiy, and T.~Brox.
\newblock Flownet 2.0: Evolution of optical flow estimation with deep networks.
\newblock In {\em Proceedings of the IEEE Conference on Computer Vision and
  Pattern Recognition}, pages 2462 -- 2470, 2017.

\bibitem{ioffe2015batch}
S.~Ioffe and C.~Szegedy.
\newblock Batch normalization: Accelerating deep network training by reducing
  internal covariate shift.
\newblock In {\em International Conference on Machine Learning}, pages
  448--456, 2015.

\bibitem{Kacorri2017}
H.~Kacorri, K.~M. Kitani, J.~P. Bigham, and C.~Asakawa.
\newblock People with visual impairment training personal object recognizers:
  Feasibility and challenges.
\newblock In {\em Proceedings of the CHI Conference on Human Factors in
  Computing Systems}, pages 5839--5849, 2017.

\bibitem{Diederik2014}
D.~P. Kingma and J.~Ba.
\newblock Adam: {A} method for stochastic optimization.
\newblock {\em CoRR}, abs/1412.6980, 2014.

\bibitem{kitani2012activity}
K.~M. Kitani, B.~D. Ziebart, J.~A. Bagnell, and M.~Hebert.
\newblock Activity forecasting.
\newblock In {\em Proceedings of the European Conference on Computer Vision},
  pages 201--214, 2012.

\bibitem{kooij2014context}
J.~F.~P. Kooij, N.~Schneider, F.~Flohr, and D.~M. Gavrila.
\newblock Context-based pedestrian path prediction.
\newblock In {\em Proceedings of the European Conference on Computer Vision},
  pages 618--633, 2014.

\bibitem{lee2017desire}
N.~Lee, W.~Choi, P.~Vernaza, C.~B. Choy, P.~H.~S. Torr, and M.~Chandraker.
\newblock Desire: Distant future prediction in dynamic scenes with interacting
  agents.
\newblock In {\em Proceedings of the IEEE Conference on Computer Vision and
  Pattern Recognition}, pages 336--345, 2017.

\bibitem{Leung2014}
T.-S. Leung and G.~Medioni.
\newblock Visual navigation aid for the blind in dynamic environments.
\newblock In {\em Proceedings of the IEEE Conference on Computer Vision and
  Pattern Recognition Workshops}, pages 153 -- 158, 2014.

\bibitem{li2013pixel}
C.~Li and K.~M. Kitani.
\newblock Pixel-level hand detection in ego-centric videos.
\newblock In {\em Proceedings of the IEEE Conference on Computer Vision and
  Pattern Recognition}, pages 3570--3577, 2013.

\bibitem{li2015delving}
Y.~Li, Z.~Ye, and J.~M. Rehg.
\newblock Delving into egocentric actions.
\newblock In {\em Proceedings of the IEEE Conference on Computer Vision and
  Pattern Recognition}, pages 287--295, 2015.

\bibitem{Ma_2016_CVPR}
M.~Ma, H.~Fan, and K.~M. Kitani.
\newblock Going deeper into first-person activity recognition.
\newblock In {\em Proceedings of the IEEE Conference on Computer Vision and
  Pattern Recognition}, pages 1894--1903, 2016.

\bibitem{Ma_2017_CVPR}
W.-C. Ma, D.-A. Huang, N.~Lee, and K.~M. Kitani.
\newblock Forecasting interactive dynamics of pedestrians with fictitious play.
\newblock In {\em Proceedings of the IEEE Conference on Computer Vision and
  Pattern Recognition}, pages 774 -- 782, 2017.

\bibitem{icml2010_NairH10}
V.~Nair and G.~E. Hinton.
\newblock Rectified linear units improve restricted boltzmann machines.
\newblock In {\em Proceedings of the International Conference on Machine
  Learning}, pages 807--814, 2010.

\bibitem{ParkCVPR16}
H.~S. Park, J.-J. Hwang, Y.~Niu, and J.~Shi.
\newblock Egocentric future localization.
\newblock In {\em Proceedings of the IEEE Conference on Computer Vision and
  Pattern Recognition}, pages 4697--4705, 2016.

\bibitem{Park2012}
H.~S. Park, E.~Jain, and Y.~Sheikh.
\newblock {3D Social Saliency from Head-mounted Cameras}.
\newblock In {\em Proceedings of the Advances in Neural Information Processing
  Systems}, pages 1--9, 2012.

\bibitem{Pirsiavash2012}
H.~Pirsiavash and D.~Ramanan.
\newblock Detecting activities of daily living in first-person camera views.
\newblock In {\em Proceedings of the IEEE Conference on Computer Vision and
  Pattern Recognition}, pages 2847--2854, 2012.

\bibitem{Poleg2014}
Y.~Poleg, C.~Arora, and S.~Peleg.
\newblock Head motion signatures from egocentric videos.
\newblock In {\em Proceedings of the Asian Conference on Computer Vision},
  pages 1--15, 2014.

\bibitem{renNIPS15fasterrcnn}
S.~Ren, K.~He, R.~Girshick, and J.~Sun.
\newblock Faster {R-CNN}: Towards real-time object detection with region
  proposal networks.
\newblock In {\em Advances in Neural Information Processing Systems}, pages
  1--9, 2015.

\bibitem{Rhinehart2017}
N.~Rhinehart and K.~M. Kitani.
\newblock First-person activity forecasting with online inverse reinforcement
  learning.
\newblock In {\em Proceedings of the IEEE International Conference on Computer
  Vision}, pages 3696--3705, 2017.

\bibitem{Robicquet2016}
A.~Robicquet, A.~Sadeghian, A.~Alahi, and S.~Savarese.
\newblock Learning social etiquette: Human trajectory understanding in crowded
  scenes.
\newblock In {\em Proceedings of the European Conference on Computer Vision},
  pages 549--565, 2016.

\bibitem{Ryoo:2015:RAP:2696454.2696462}
M.~S. Ryoo, T.~J. Fuchs, L.~Xia, J.~Aggarwal, and L.~Matthies.
\newblock Robot-centric activity prediction from first-person videos: What will
  they do to me?
\newblock In {\em Proceedings of the ACM/IEEE International Conference on
  Human-Robot Interaction}, pages 295--302, 2015.

\bibitem{Ryoo_2013_CVPR}
M.~S. Ryoo and L.~Matthies.
\newblock First-person activity recognition: What are they doing to me?
\newblock In {\em Proceedings of the IEEE Conference on Computer Vision and
  Pattern Recognition}, pages 2730--2737, 2013.

\bibitem{Saran2015}
A.~Saran, D.~Teney, and K.~M. Kitani.
\newblock Hand parsing for fine-grained recognition of human grasps in
  monocular images.
\newblock In {\em Proceedings of the IEEE/RSJ International Conference on
  Intelligent Robots and Systems}, pages 1--7, 2015.

\bibitem{Schneider2013}
N.~Schneider and D.~M. Gavrila.
\newblock Pedestrian path prediction with recursive bayesian filters: A
  comparative study.
\newblock In {\em Proceedings of the German Conference on Pattern Recognition},
  pages 174--183, 2013.

\bibitem{Su_2017_CVPR}
S.~Su, J.~Pyo~Hong, J.~Shi, and H.~Soo~Park.
\newblock Predicting behaviors of basketball players from first person videos.
\newblock In {\em Proceedings of the IEEE Conference on Computer Vision and
  Pattern Recognition}, pages 1501--1510, 2017.

\bibitem{Tang2014}
T.~J.~J. Tang and W.~H. Li.
\newblock An assistive eyewear prototype that interactively converts 3d object
  locations into spatial audio.
\newblock In {\em Proceedings of the ACM International Symposium on Wearable
  Computers}, pages 119--126, 2014.

\bibitem{Tian2013}
Y.~Tian, Y.~Liu, and J.~Tan.
\newblock Wearable navigation system for the blind people in dynamic
  environments.
\newblock In {\em Proceedings of the Cyber Technology in Automation, Control
  and Intelligent Systems}, pages 153 -- 158, 2013.

\bibitem{chainer_learningsys2015}
S.~Tokui, K.~Oono, S.~Hido, and J.~Clayton.
\newblock Chainer: a next-generation open source framework for deep learning.
\newblock In {\em Proceedings of Workshop on Machine Learning Systems}, pages
  1--6, 2015.

\bibitem{Xie2013}
D.~Xie, S.~Todorovic, and S.~C. Zhu.
\newblock Inferring dark matter and dark energy from videos.
\newblock In {\em Proceedings of the IEEE International Conference on Computer
  Vision}, pages 2224--2231, 2013.

\bibitem{Ye2015}
Z.~Ye, Y.~Li, Y.~Liu, C.~Bridges, A.~Rozga, and J.~M. Rehg.
\newblock Detecting bids for eye contact using a wearable camera.
\newblock In {\em Proceedings of the IEEE International Conference on Automatic
  Face and Gesture Recognition}, pages 1--8, 2015.

\bibitem{yi2016pedestrian}
S.~Yi, H.~Li, and X.~Wang.
\newblock Pedestrian behavior understanding and prediction with deep neural
  networks.
\newblock In {\em Proceedings of the European Conference on Computer Vision},
  pages 263--279, 2016.

\bibitem{yonetani2016recognizing}
R.~Yonetani, K.~M. Kitani, and Y.~Sato.
\newblock Recognizing micro-actions and reactions from paired egocentric
  videos.
\newblock In {\em Proceedings of the IEEE Conference on Computer Vision and
  Pattern Recognition}, pages 2629--2638, 2016.

\bibitem{Zhang_2017_CVPR}
M.~Zhang, K.~Teck~Ma, J.~Hwee~Lim, Q.~Zhao, and J.~Feng.
\newblock Deep future gaze: Gaze anticipation on egocentric videos using
  adversarial networks.
\newblock In {\em Proceedings of the IEEE International Conference on Computer
  Vision}, 2017.

\bibitem{zhou2017unsupervised}
T.~Zhou, M.~Brown, N.~Snavely, and D.~G. Lowe.
\newblock Unsupervised learning of depth and ego-motion from video.
\newblock In {\em Proceedings of the IEEE Conference on Computer Vision and
  Pattern Recognition}, pages 1851 -- 1860, 2017.

\end{thebibliography}

}

\end{document}